\documentclass[twocolumn]{article}
\usepackage{authblk}

\usepackage[english]{babel}
\usepackage[utf8x]{inputenc}
\usepackage[T1]{fontenc}

\usepackage[letterpaper,top=0.75in,bottom=0.75in,left=0.5in,right=0.5in]{geometry}
\usepackage{parskip} 

\usepackage{amsmath}
\usepackage{graphicx}
\usepackage[colorlinks=true, allcolors=blue]{hyperref}
\usepackage{tabularx}
\usepackage{ragged2e} % for better text alignment
\usepackage{booktabs} % for better horizontal rules
\usepackage{comment}
\usepackage{pgffor}
\usepackage{subcaption}
\usepackage{enumitem}
\usepackage{multirow}

\usepackage[backend=biber,sorting=none,style=numeric]{biblatex}

\usepackage{siunitx}
\usepackage[normalem]{ulem}

\usepackage{etoolbox}
\AtBeginEnvironment{tabularx}{\footnotesize}

\def\SIMPLE{}

\newcommand{\papertitle}{LookAroundNet: Extending Temporal Context with Transformers for Clinically Viable EEG Seizure Detection}

\title{\papertitle}
\author[1]{Þór Sverrisson}
\author[1]{Steinn Guðmundsson}
\affil[1]{University of Iceland, Faculty of Computer Science}

\begin{document}

\maketitle

\begin{abstract}
Automated seizure detection from electroencephalography (EEG) remains difficult due to the large variability of seizure dynamics across patients, recording conditions, and clinical settings. We introduce LookAroundNet, a transformer-based seizure detector that uses a wider temporal window of EEG data to model seizure activity. The seizure detector incorporates EEG signals before and after the segment of interest, reflecting how clinicians use surrounding context when interpreting EEG recordings. We evaluate the proposed method on multiple EEG datasets spanning diverse clinical environments, patient populations, and recording modalities, including routine clinical EEG and long-term ambulatory recordings, in order to study performance across varying data distributions. The evaluation includes publicly available datasets as well as a large proprietary collection of home EEG recordings, providing complementary views of controlled clinical data and unconstrained home-monitoring conditions. Our results show that LookAroundNet achieves strong performance across datasets, generalizes well to previously unseen recording conditions, and operates with computational costs compatible with real-world clinical deployment. The results indicate that extended temporal context, increased training data diversity, and model ensembling are key factors for improving performance. This work contributes to moving automatic seizure detection models toward clinically viable solutions.
\end{abstract}

\section{Introduction}

Epilepsy is a chronic brain disorder that causes unprovoked seizures due to temporary electrical disturbance in the brain. It affects people of all ages worldwide and arises from various causes, including head trauma, infections, and genetic and environmental factors. However, in many cases, the underlying cause remains unknown, and symptoms often vary considerably between individuals. Epilepsy can greatly affect quality of life, but approximately three-quarters of those diagnosed can manage seizures with medication or surgery~\cite{duncan2006adult, giourou2015introduction, schmidt2014drug, thijs2019epilepsy}. Effective treatment depends on an accurate diagnosis, which is typically based on the detection and analysis of seizures in the electroencephalogram (EEG) signal, a non-invasive recording of the brain's electrical activity captured through scalp electrodes~\cite{smith2005eeg}.

An EEG recording can vary noticeably depending on the patient, setting and duration of the monitoring~\cite{tatum2018clinical}. Age plays a significant role, with neonatal, pediatric, and adult EEGs often differing substantially in both appearance and interpretation~\cite{eisermann2013normal, hashemi2016characterizing}. Clinical recordings, such as routine EEGs, are performed in a medical setting and monitor the patient for a short period in a controlled environment. In contrast, ambulatory EEGs are often performed at home, providing extended monitoring over several days while the patient continues daily activities, increasing the chances of detecting infrequent epileptic events missed during short clinical visits~\cite{narayanan2008latency}. 

Proper diagnosis often requires long-term monitoring, leading to increased use of home-based video-EEG, combining EEG recordings with synchronized video footage of the patient, which reduces the burden on both patients and healthcare systems~\cite{tatum2021ambulatory,benbadis2020role}. However, ambulatory recordings are usually more prone to artifacts, such as muscle movement and electrode displacement, and applying artifact removal techniques in this setting can be challenging~\cite{minguillon2017trends}.

EEGs are typically reviewed through visual examination of a montage, which arranges EEG channels to display brain activity across the scalp. The review process is labor-intensive and time-consuming, requiring expert knowledge and careful analysis to accurately identify seizures from the complex patterns in the EEG signal. Accurate identification often relies on examining the surrounding context, reviewing EEG segments before and after a suspected event to observe gradual changes or sudden shifts that may indicate seizure activity~\cite{koutroumanidis2017role, fisher2014can}. Variability in seizure appearance across individuals, recording conditions, and equipment further complicates detection~\cite{amin2023normal, hairston2014usability}.

These challenges have spurred growing interest in automated seizure detection methods that use algorithms to efficiently analyze EEG data. Such methods can be classified into real-time and offline approaches, with real-time systems providing immediate alerts for prompt intervention, whereas offline systems can reduce clinicians' workload and expedite the diagnostic process~\cite{baumgartner2018seizure}. Over time, research in seizure detection has transitioned from rule-based techniques to the widespread adoption of deep neural networks~\cite{shoeibi2021epileptic}. These models have demonstrated strong capabilities in analyzing complex patterns, enabling their application across various healthcare domains~\cite{esteva2019guide}. However, despite advancements, automated seizure detection methods still lack the precision and reliability of human specialists~\cite{reus2022automated}. 

The performance of deep learning models depends heavily on both the amount of training data and the quality of the labels in supervised settings. Therefore, the limited availability of publicly accessible EEG seizure datasets, each with distinct characteristics, poses challenges for developing models that can generalize effectively across different data sources~\cite{wong2023eegB}. Curating these datasets is resource-intensive and costly, and their distribution is often constrained by strict regulations surrounding personal health information~\cite{white2022data}. Furthermore, existing datasets are typically small in size, often representing recordings from just a handful of patients. This lack of diversity can lead to substantial variability in data distributions, resulting in models trained on a particular dataset facing difficulties when generalizing to more diverse populations~\cite{wong2023eegB}. Moreover, disagreements between annotators on the labeling and boundaries of seizure events can introduce inconsistencies and impact model training~\cite{borovac2021influence, swisher2015diagnostic}.

Evaluating seizure detection models presents additional challenges due to varying application needs and the lack of standardized validation protocols~\cite{dan2025szcore}. In some cases, detecting the exact onset of a seizure is crucial, particularly for real-time interventions. In other cases, identifying the presence of a seizure is more important than detecting its exact location, such as when assessing the effectiveness of pharmacological treatment. These differences lead to varying validation methods and performance metrics across studies, making it difficult to compare results directly. Furthermore, reported performance metrics often do not reflect clinical translatability, as evaluations may omit considerations such as model generalisability, run-time constraints, explainability, and clinically relevant metrics~\cite{moutonnet2024clinical}.

In this paper, we present a transformer-based model architecture and conduct a comprehensive evaluation across several independent EEG datasets, which vary in clinical settings and patient demographics. The proposed model incorporates surrounding temporal context when classifying EEG segments, and we evaluate the impact of combining training data from multiple sources to enhance generalization. We employ the SzCORE~\cite{dan2025szcore} framework for model evaluation and assess the model's performance on both publicly available and private datasets. In particular, we focus on a large proprietary dataset consisting of ambulatory home recordings to evaluate its clinical relevance.

\section{Related Work}

Recent research in automatic seizure detection increasingly favors deep learning–based approaches, particularly Convolutional Neural Networks (CNNs) and Transformer architectures. Most CNN-based studies employ a sliding-window strategy, dividing continuous EEG recordings into shorter segments that are processed independently~\cite{thodoroff2016learning,o2020neonatal,wang2021one}. This allows CNNs to effectively capture spatial patterns and short-term temporal dynamics. For instance, Thuwajit et al.~\cite{thuwajit2021eegwavenet} used depthwise convolutions to extract channel-specific features from 4-second EEG segments, followed by spatiotemporal convolutions to model inter-channel dependencies. However, fixed window lengths limit the model’s temporal awareness, making it difficult to capture the progressive development of seizure activity.

In contrast, Chatzichristos et al.~\cite{chatzichristos2020epileptic} introduced a CNN composed of three attention-gated U-Nets, each processing a different view of the EEG signal derived from distinct pre-processing strategies. Their model analyzes entire recordings in a single pass, generating predictions for every time point. Seeuws et
al.~\cite{seeuws2024avoiding} extended this idea using an event-based training objective that predicts seizure centers and durations. Whole-recording approaches capture long-term temporal dependencies but are computationally intensive for extended EEG monitoring and assume full access to the recording, which limits their suitability for real-time applications.

Effectively capturing temporal context beyond short, localized windows remains a key challenge in EEG analysis. Capturing longer-term dependencies is important to help separate abnormal patterns from background activity, an insight that dates back to the earliest efforts to develop automated seizure detectors~\cite{gotman1982automatic}. Early approaches, such as Temko et al.~\cite{temko2013robust}, proposed averaging the preceding 10 minutes of non-seizure EEG activity to dynamically adjust the seizure probability threshold. Xun et al.~\cite{xun2016detecting} proposed a Continuous Bag-of-Words–inspired approach to extract temporal features through context learning. Subsequent work has sought to extend temporal modeling capacity through alternative model architectures~\cite{hussein2018epileptic, golmohammadi2018deep}.

Transformer models~\cite{vaswani2017attention} have gained traction in EEG analysis due to their self-attention mechanism, which inherently models long-range relationships. Many studies combine convolutional front-ends for local feature extraction with transformer modules to model broader temporal context~\cite{benfenati2024biseizure, zhu2023automated}. Wu et al.~\cite{wu2025seizuretransformer} achieved good results by extending the U-Net framework with a transformer encoder, using a 60-second window to generate sample-level predictions. Additionally, studies have shown that transformers can scale to very large EEG models with substantial representational capacity~\cite{jiang2024large, wang2024cbramod}. However, this scalability can introduce significant computational demands during both training and deployment.

Although many studies report strong performance, most evaluate their models on a single dataset, which limits the generalizability of their findings. Studies that evaluate on multiple datasets have sometimes demonstrated limited generalization across out-of-sample datasets~\cite{dan2025szcore}. These challenges motivate our interest in two core aspects: (1) effectively modeling the temporal context surrounding EEG segments of interest and (2) achieving robust generalization across datasets alongside performance that remains clinically reliable. We believe both aspects are essential for models intended for real-world deployment. Guided by these considerations, this study seeks to address these challenges using transformer-based models.

\section{Methods}

\subsection{Data}
The datasets used in this study are summarized in Table~\ref{tab:datasets}, with additional details provided in Appendix~\ref{app:dataset}. These datasets contain EEG recordings collected from multiple sources and captured in various environments, with recordings ranging from a few minutes to several days of monitoring. All recordings were conducted using the standard 10-20 electrode system and include seizure annotations.

\begin{table}[t]
\caption{Datasets used in the study, grouped by training, validation, and test splits.}
\label{tab:datasets}
\centering

\begin{tabularx}{\columnwidth}{X r r r r}
\toprule
Dataset & Patients (w/ sz) & \# Seizures & Sz Ratio (\%) & Duration \\
\midrule
\multicolumn{5}{l}{\textit{Training}} \\
\midrule
TUSZ & 579 (208) & 2421 & 5.34 & 37d 22h \\
Kvikna$^{\dagger}$ & 254 (254) & 1099 & 0.09 & 589d 07h \\
\addlinespace
\multicolumn{5}{l}{\textit{Validation}} \\
\midrule
TUSZ & 53 (45) & 1081 & 4.58 & 18d 03h \\
Kvikna$^{\dagger}$ & 36 (36) & 243 & 0.16 & 66d 07h \\
\addlinespace
\multicolumn{5}{l}{\textit{Test}} \\
\midrule
TUSZ & 43 (34) & 469 & 5.93 & 05d 07h \\
Kvikna$^{\dagger}$ & 49 (49) & 514 & 0.33 & 132d 13h \\
Siena & 14 (14) & 47 & 0.56 & 05d 23h \\
SeizeIT1 & 42 (42) & 182 & 0.08 & 175d 11h \\
\bottomrule
\end{tabularx}

\vspace{2pt}
\parbox{\columnwidth}{%
Each dataset lists the number of patients (with seizures), seizures, total recording duration, and the seizure-to-total duration ratio.
$^{\dagger}$The Kvikna dataset is proprietary; all other datasets are public.
}
\end{table}

\subsubsection{TUSZ}

The Temple University Hospital EEG Seizure Corpus (TUSZ)~\cite{shah2018temple} is a dataset created to support research in automatic seizure detection. TUSZ is a subset of a comprehensive collection of clinical EEG recordings from various hospital settings, including ICU and epilepsy monitoring units. Seizures were annotated by trained undergraduates from Temple University through a collaborative review process based on EEG data.

\subsubsection{SeizeIT1}

SeizeIT1~\cite{chatzi2023seizeit1} includes EEG recordings from patients with refractory epilepsy who underwent presurgical evaluation at the University Hospitals Leuven in Belgium. Patients were monitored in a home environment over multiple days and were restricted to move within their room. Seizures were annotated by a neurologist based on video EEG assessments. For 12 seizures, no end time was annotated. For these, we assign a fixed duration of 30 seconds. This matches the approach used in the SzCORE benchmark and ensures our results are directly comparable.

\subsubsection{Siena}

The Siena Scalp EEG Database~\cite{detti2020siena} includes EEG recordings collected by the Unit of Neurology and Neurophysiology at the University of Siena, Italy. During the recordings, patients were instructed to remain in bed as much as possible, either asleep or awake. Seizure annotations were performed by expert clinicians using video EEG.

\subsubsection{Kvikna}

The Kvikna EEG Seizure Corpus v1 is a proprietary subset of ambulatory EEG recordings from Kvikna Medical's research database. The recordings were collected from patients with suspected epilepsy, who were monitored at home over several days. Patients were permitted to move freely within their homes but were instructed to remain within a camera’s view for most of the time. Two cameras were installed in each home, one in the bedroom and one in the main daytime area. All selected recordings were reviewed and labeled by experts using video EEG.

\subsection{Implementation}

\subsubsection{Model Architecture}

\begin{figure}[!t]
\centerline{\includegraphics[width=\columnwidth]{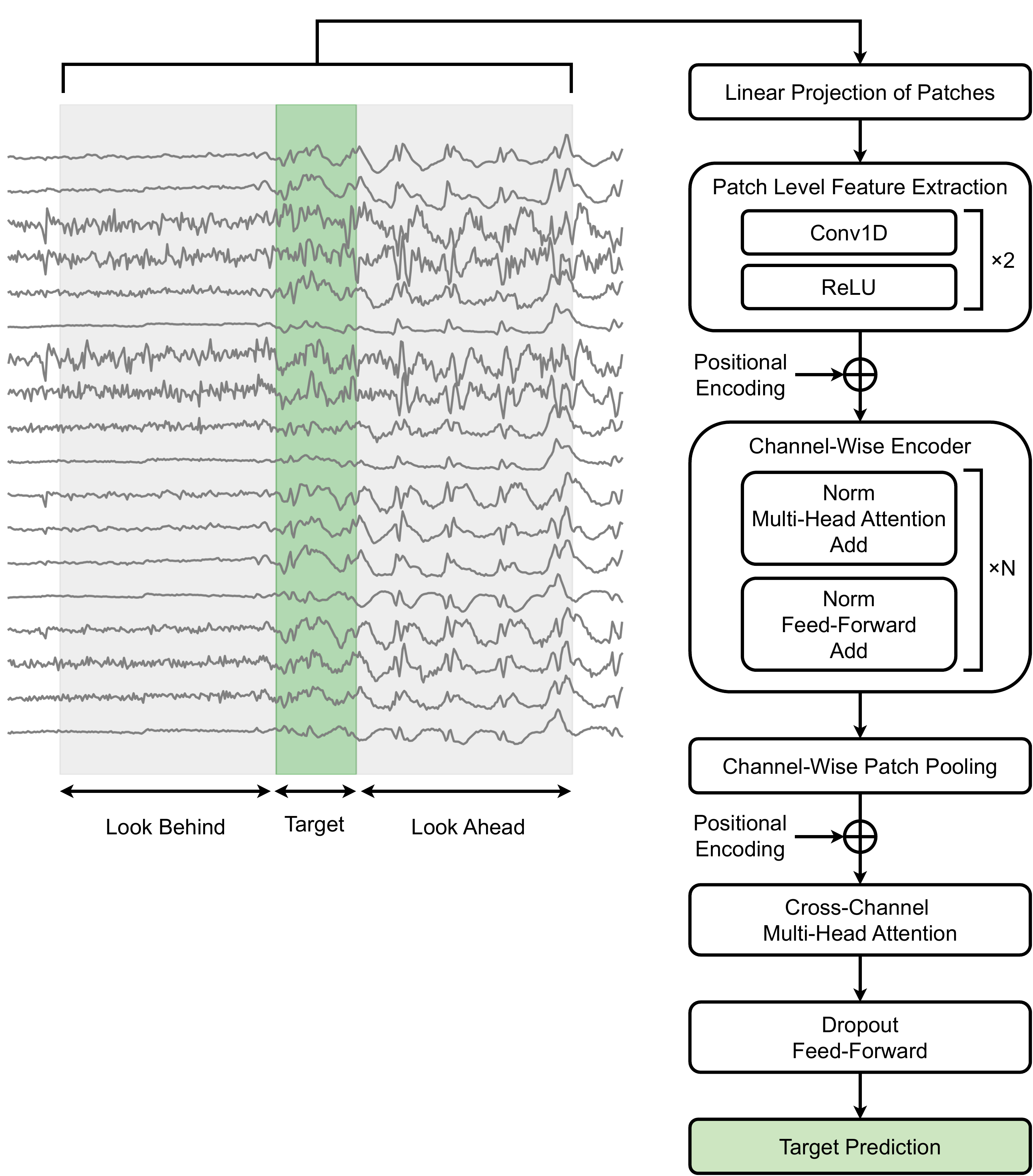}}
\caption{The LookAroundNet architecture. The input window includes a target segment, which is the specific EEG segment to be classified, along with surrounding context segments that provide additional temporal information. Each input channel is independently patched, and features are extracted from each patch. A transformer encoder processes each channel, followed by multi-head attention across channels and a classification layer.}
\label{fig:transformer}
\end{figure}

Fig.~\ref{fig:transformer} illustrates the architecture of the proposed LookAroundNet transformer. The model takes as input a continuous, multi-channel EEG segment divided into three parts: a central target section, a look-behind section, and a look-ahead section. The look-behind and look-ahead sections together form the context window, which provides temporal context from both past and future signal activity relative to the target segment. This contextual information enables the model to capture temporal dependencies and dynamic patterns that extend beyond the immediate target region. This design mirrors the approach of human reviewers, who interpret EEG signals by integrating observations of preceding and subsequent EEG activity.

The architectural design of LookAroundNet reflects a common pattern in seizure detection models, where temporal dynamics are first extracted from individual EEG channels, followed by spatial modeling to capture inter-channel relationships. The input is first divided into non-overlapping temporal patches, where each patch is a fixed-duration slice from a single channel, mapped via linear projection. We apply a two-layer convolutional feature extractor to the patches. We experimented with both fewer and more layers, but fewer layers noticeably reduced performance, whereas additional layers provided no noticeable improvement.

The patch embeddings are then processed by channel-wise transformer encoder layers, with the temporal order of the patches preserved through learnable positional encodings. Mean pooling is then applied across the patch dimension, resulting in a single aggregated representation for each channel. A multi-head attention mechanism with learnable positional encodings is applied across channels to capture inter-channel interactions. The resulting representations are passed through a fully connected layer to produce the final class prediction for the target segment.

\subsubsection{Data Preparation}

The raw EEG signal undergoes pre-processing to standardize its format and improve uniformity across recordings, addressing variations from different settings, sampling rates, and electrode references. First, the signal is transformed into a Longitudinal Bipolar Montage\footnote{The Longitudinal Bipolar Montage used has the following channels: Fp2-F4, F4-C4, C4-P4, P4-O2, Fp1-F3, F3-C3, C3-P3, P3-O1, Fp2-F8, F8-T4, T4-T6, T6-O2, Fp1-F7, F7-T3, T3-T5, T5-O1, Fz-Cz, Cz-Pz.}. For electrodes included in the montage but missing in the raw recording, their values are estimated by averaging signals from the two spatially closest channels. The signal is then filtered and resampled using the MNE library~\cite{GramfortEtAl2013a}. This includes FIR filtering with a 0.5 Hz high-pass filter to remove low-frequency drift, a 64 Hz low-pass filter to reduce high-frequency noise, and a 50 or 60 Hz notch filter to eliminate power line interference. Reflective padding is applied at the signal boundaries to minimize edge artifacts from filtering. Finally, the signal is downsampled to 128 Hz.

\subsubsection{Experimental Setup}

In our experiments, we pre-process the complete datasets before training and evaluation. This setup reflects an offline analysis scenario, as opposed to online processing where smaller data segments would be handled incrementally. In each training epoch, 60k segments are uniformly sampled from the full training set across three categories: fully seizure segments, defined as those with continuous seizure activity across the entire target section; fully non-seizure segments, with no seizure activity in the target section; and mixed segments, where the target section includes both seizure and non-seizure activity. The target segment duration is fixed to 16 seconds for all experiments. To minimize bias toward individual subjects, we sample evenly across patients in the training set, and within each patient, we sample equally from continuous seizure, non-seizure, and mixed portions to balance individual seizure events regardless of their duration.

All models are trained for 200 epochs, and in each epoch, segments are sampled from the full dataset with randomly selected start times. The Kvikna training set comprises very long recordings, spanning almost 600 days. Confidence in the annotations in the portions surrounding marked seizures in this set is high, but there is less confidence in other portions of the recordings. We therefore divide each recording in the Kvikna training set into non-overlapping hour-long segments and retain only those containing at least one seizure for training, discarding the rest. This approach also reduces computational cost and avoids severe class imbalance.

During optimization, segments are assigned a binary label, seizure or non-seizure. For mixed segments, the label is determined by the predominant activity within the target section, defined by whether seizure activity exceeds half of the duration. On the combined TUSZ and Kvikna validation sets, we found that a prediction threshold of 0.5 was suboptimal, as higher thresholds improved the F1-score and reduced false positives, thereby making the model more suitable for clinical purposes. We therefore fixed the threshold at 0.85, and the model achieving the highest validation F1-score at this threshold was selected during training. This model–threshold combination was then used for testing.

All experiments are conducted using four NVIDIA Quadro RTX 5000 GPUs, except for inference time measurements, which use different hardware as described in Section~\ref{sec:inference}. The models are trained with CrossEntropyLoss and the AdamW Schedule-Free optimizer~\cite{defazio2024road}, using the default parameters. We found that the schedule-free setup stabilizes optimization and prevents divergence during training. We use a cross-entropy loss function with label smoothing, which in our experiments reduced the occurrence of loss spikes during training. We also experimented with gradient clipping, but it resulted in worse performance (data not shown). A complete overview of the model’s architectural and training hyperparameters can be found in Appendix~\ref{app:modelarchitecture}.

\subsubsection{Evaluation}

We evaluate the performance of our model across different test sets using the SzCORE framework~\cite{dan2025szcore}, which is specifically designed to standardize the validation of seizure detection methods. It introduces two primary scoring methods: sample-based and event-based scoring. Sample-based scoring compares predicted and reference labels at a fixed frequency to assess detection precision at the sample level. In contrast, event-based scoring focuses on the overlap between predicted and reference seizure events, providing a more clinically relevant measure. We follow SzCORE recommendations in the choice of evaluation metrics, using sensitivity, F1-score, precision, and false predictions per day (FP/day). 

During inference, we iterate over the EEG signal in overlapping segments, advancing the window by 2 seconds at a time. This choice balances computational efficiency with temporal resolution, and predictions from overlapping segments are averaged to generate a continuous binary output. For evaluation, the model output is compared to the ground-truth labels at a frequency of 1 Hz. All reported results are averaged over 5 independently trained models.

\section{Results and Discussion}

\subsection{Model Performance}

\begin{table*}[t]
\caption{SzCORE performance metrics for different models across datasets and evaluation methods.}
\label{tab:performance}
\centering

\begin{tabularx}{\textwidth}{
    X  % Model
    r % # Params (M)
    l % Metrics
    r r r r                         % Event-based (4)
    r r r r                         % Sample-based (4)
}
\toprule
            \multirow{2}{*}{Model} &
            \multirow{2}{*}{\# Params (M)} &
            \multirow{2}{*}{Metrics} & 
            \multicolumn{4}{c}{Event based} & \multicolumn{4}{c}{Sample based} \\
            \cmidrule(lr){4-7} \cmidrule(lr){8-11}
             & & & {TUSZ} & {Siena} & {SeizeIT} & {Kvikna} & {TUSZ} & {Siena} & {SeizeIT} & {Kvikna}\\
            \midrule
            \multirow{4}{*}{\shortstack{EventNet~\cite{seeuws2024avoiding}}} 
             & \multirow{4}{*}{1.7} 
             & F1-score    & 59.2 & 51.7 & 22.1 &     & 44.5 & 43.7 & 16.8 &     \\
             & & Sensitivity & 53.9 & 58.7 & 61.9 &     & 44.2 & 53.2 & 51.6 &     \\
             & & Precision   & 43.2 & 53.7 & 16.7 &     & 33.3 & 45.5 & 13.1 &     \\
             & & FP/day      & 28.4 &  8.5 & 19.6 &     &      &      &      &     \\
            \midrule
            \multirow{4}{*}{\shortstack{EEG-U-Transformer~\cite{wu2025seizuretransformer}}} 
             & \multirow{4}{*}{41.0} 
             & F1-score    & 63.6 &      & 45.5 &     & 40.8 &     & 28.2 &     \\
             & & Sensitivity & 57.3 &      & 37.5 &     & 29.3 &     & 16.1 &     \\
             & & Precision   & 49.7 &      & 56.2 &     & 50.1 &     & 63.7 &     \\
             & & FP/day      & 19.3 &      & 0.9  &     &      &     &      &     \\
            \midrule
            \multirow{4}{*}{\shortstack{LookAroundNet {\scriptsize $(-32, 32)$} \\ (This study)}}
             & \multirow{4}{*}{0.5} 
             & F1-score    & 72.1 & 54.9 & 26.4 & 21.8 & \underline{\num{62.1}} & 49.0 & 31.0 & 17.6\\
             & & Sensitivity & 74.2 & 69.1 & 58.5 & 47.4 & 51.9 & 42.0 & 29.3 & 16.4\\
             & & Precision   & 70.3 & 46.0 & 17.1 & 14.2 & 77.4 & 59.4 & 33.0 & 18.9\\
             & & FP/day      & 12.7 & 3.2  & 1.7  & 6.5  &      &      &      &  \\
             \midrule
             \midrule
            \multirow{4}{*}{\shortstack{LookAroundNet {\scriptsize (ensemble)} \\ (This study)}}
              & \multirow{4}{*}{1.4} 
              & F1-score       & \underline{\num{77.8}} & \underline{\num{68.5}} & \underline{\num{47.0}}  & \underline{\num{31.2}} & 61.8 & \underline{\num{49.9}} & \underline{\num{37.3}} & \underline{\num{19.9}} \\
              & & Sensitivity  & 69.2 & 62.7 & 49.0 & 38.4 & 46.7  & 37.5 & 25.4  & 13.3  \\
              & & Precision    & 89.0  & 75.6  & 46.2  & 26.6 & 91.6  & 74.5  & 71.0  & 40.2 \\
              & & FP/day       & 3.4  & 0.8  & 0.4 & 2.5  &  &  & &  \\
            \bottomrule
                
        \end{tabularx}

\vspace{2pt}
\parbox{\textwidth}{%
The numbers for EventNet and EEG-U-Transformer are obtained from the SzCORE benchmark results~\cite{dan2025}.
Highest F1-scores for each test set are underlined.%
}
\end{table*}

The results, along with a comparison with selected models from the SzCore benchmark~\cite{dan2025}, are presented in Table~\ref{tab:performance}. The comparison considers models that performed favorably on the benchmark, including EEG-U-Transformer~\cite{wu2025seizuretransformer}, the winner of the 2025 Seizure Detection Challenge, and EventNet~\cite{seeuws2024avoiding}. EventNet is trained on TUSZ, EEG-U-Transformer on TUSZ and Siena, and LookAroundNet uses TUSZ and Kvikna. The LookAroundNet results are reported for two configurations: a single-model setup using a 16-second target window with a 32-second look-behind and a 32-second look-ahead, and an ensemble variant described in Section~\ref{sec:Ensembling}. SzCORE values for all experiments are provided in Appendix~\ref{app:completeresults}.

When comparing the single-model variant of LookAroundNet to prior models, the model consistently outperforms existing work on TUSZ and shows overall stronger performance on both Siena and SeizeIT1. Despite having significantly fewer parameters than competing architectures, it achieves higher F1-scores for all test sets, except for the event-based measure on SeizeIT1. Although its performance on this metric is lower than that of EEG-U-Transformer, LookAroundNet compensates with substantially higher event-based sensitivity, detecting many more seizures at the cost of less than one additional false detection per day.

It is important to note that F1-scores do not fully capture the overall performance of a model or its practical applicability. This metric is particularly sensitive to the balance between seizure and non-seizure data, which leads to considerable variation in F1-scores across the datasets in our study. An often more clinically relevant view is to consider the balance between the seizure detection rate (SDR) and the false positive rate. LookAroundNet achieves a strong trade-off between these measures on the Siena and SeizeIT datasets. Performance on Kvikna is somewhat lower, and the false positives per day (FP/day) are comparatively higher for TUSZ.

The reduced performance on the Kvikna dataset likely stems from its composition of long-term home recordings, where the lack of a controlled environment increases the presence of artifacts and signal degradation. Closer inspection revealed that a small number of recordings disproportionately contribute to the false positive rate, primarily due to significant EEG artifacts. These false positives often cluster within segments where signal quality deteriorates, typically caused by loose electrodes or other non-neural disturbances.

Further inspection of the Kvikna dataset revealed that some false positive events were, in fact, genuine seizures that had been missed during the labeling process. No notable differences in performance were observed with respect to the time of day or patient demographics. However, using 16-second segments causes seizures shorter than 8 seconds to go unlabeled, as the classifier requires more than half of a segment to contain seizure activity. In the TUSZ test set, roughly 15\% of seizures fall below this threshold. A visualization of the model output for all test sets is provided in Appendix~\ref{app:visualoutput}.

\subsection{Impact of Training Set}

\begin{figure*}[t]
\centerline{\includegraphics[width=\textwidth]{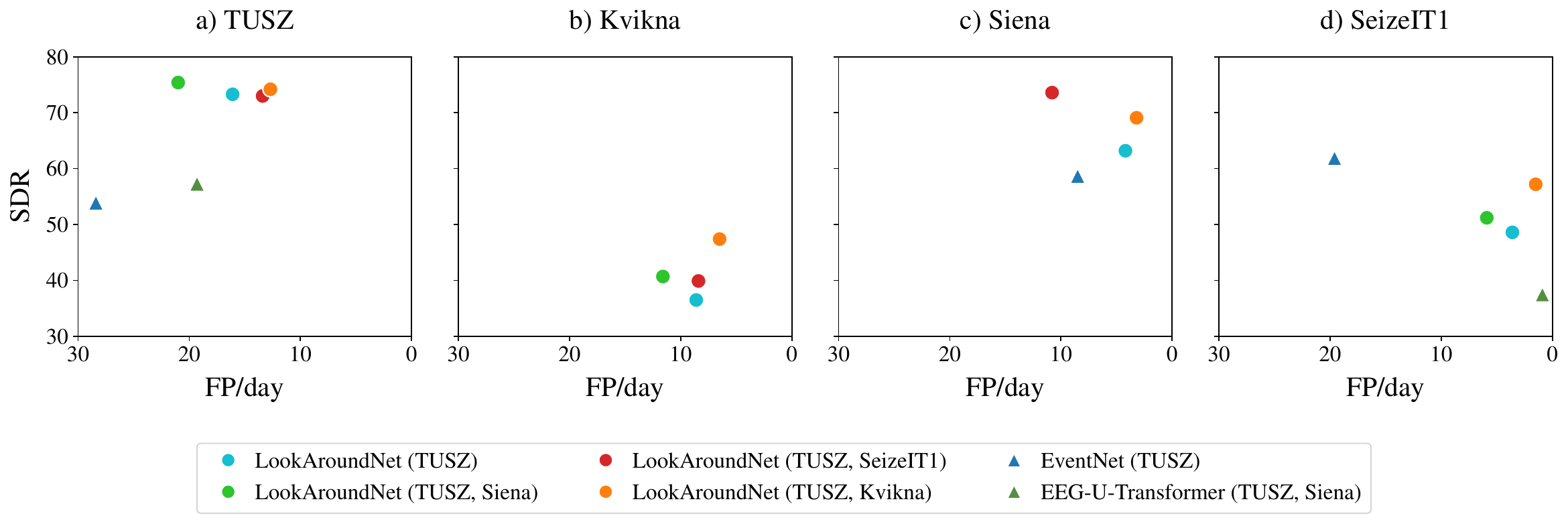}}
\caption{Test set performance of selected models and training set combinations, shown as seizure detection rate versus false positives per day. Each subfigure corresponds to a different test set, with points representing individual model–training set combinations.}
\label{fig:trainingsets}
\end{figure*}

To study the effect of training data and to compare our results with previous work, we train LookAroundNet on different combinations of public datasets and compare the results to those obtained when the Kvikna dataset is included (see Fig.~\ref{fig:trainingsets}). We examine the seizure detection rate, given by the event-based sensitivity, relative to false positives per day.

For LookAroundNet, we train on TUSZ alone and on TUSZ combined with Siena, SeizeIT1, or Kvikna. Across these experiments, combining TUSZ with Kvikna consistently gave the best results, producing a model that generalized relatively well across all test datasets. In contrast, adding Siena improved the SDR but substantially increased the number of false positives per day. Adding SeizeIT1 slightly improved performance on TUSZ and Kvikna but caused a notable increase in false detections on Siena. The limited benefit observed from incorporating the Siena or SeizeIT1 datasets into the training set likely reflects their relatively small contribution of additional seizures and patients compared with the substantially larger TUSZ and Kvikna datasets.

For a fair comparison to previous studies, we compare LookAroundNet when limited by the same publicly available datasets used in those studies. When trained only on the TUSZ dataset, LookAroundNet achieves a significantly higher seizure detection rate and fewer false positives on both TUSZ and Siena, compared to EventNet and EEG-U-Transformer. On SeizeIT1, however, the relative ranking of models is less definitive when considering the trade-off between detection rate and false positives per day. Nonetheless, these results suggest that LookAroundNet’s design and training approach help achieve consistent performance across datasets.

\subsection{Context Window Variation}

\begin{figure*}[t]
\centerline{\includegraphics[width=\textwidth]{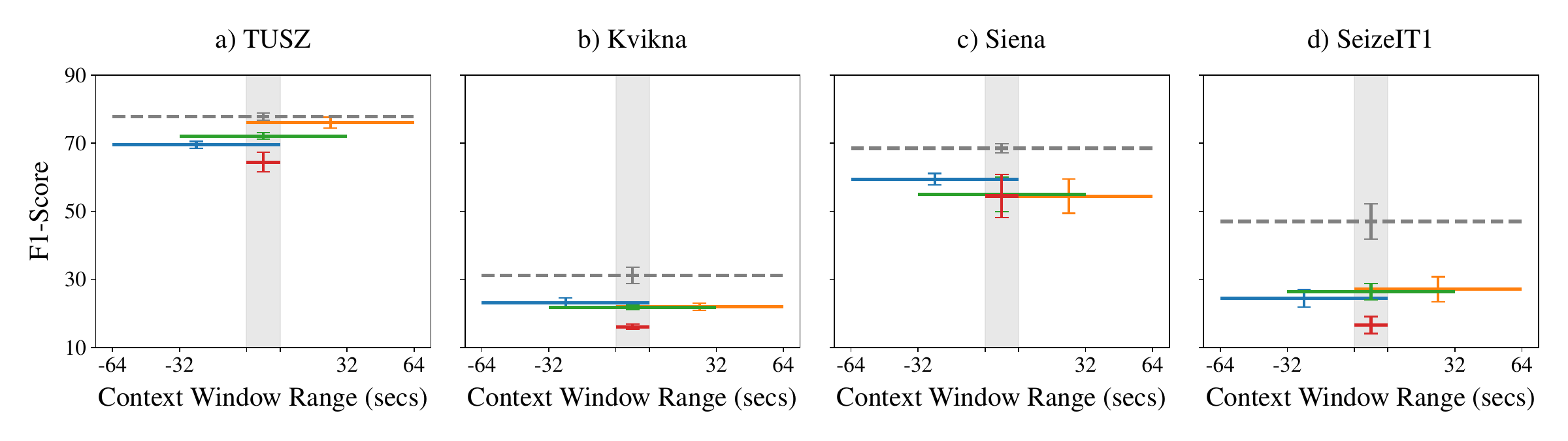}}
\caption{Comparison of event-based F1-scores across test datasets for different context window configurations, each with a total extended context duration of 64 seconds. The highlighted area corresponds to the target segment, and the size and positions of the colored bars represent the context window size and placement. A negative context size indicates past context, whereas a positive size indicates future context. The dashed line represents the ensembled model, created by combining outputs from the three look-around configurations.}
\label{fig:windowvariation}
\end{figure*}

We experimented with shifting the position of the context window relative to the prediction target and varying its duration. Fig.~\ref{fig:windowvariation} compares model performance when the context window is placed entirely before, centered on, or entirely after the target segment. We report the corresponding event-based F1-scores for each configuration and include a baseline model trained without any surrounding context for reference.

Providing surrounding context consistently improved performance compared to the no-context baseline. This highlights the value of allowing the model to look around the target segment. However, comparing F1-scores across different context window placements shows no consistent advantage for any particular placement. It's worth noting that incorporating future context introduces a detection delay equal to its duration, since the model must wait for upcoming samples before generating an output. In offline analysis, this delay is not relevant since the full recording is available at any time. For real-time applications, however, it can be preferable to rely on past context to minimize latency.

\begin{figure*}[t]
\centerline{\includegraphics[width=\textwidth]{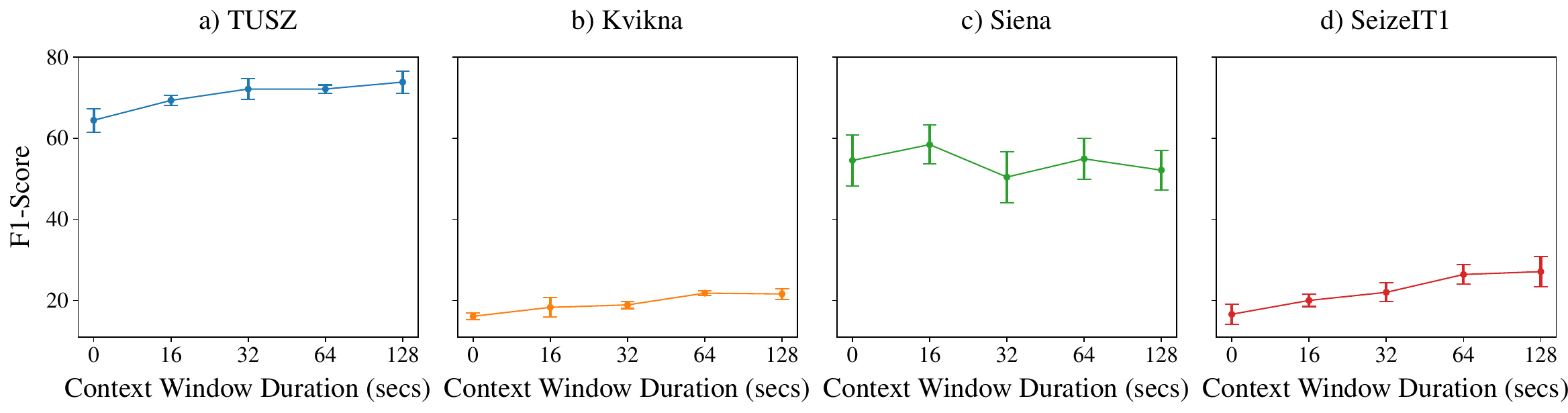}}
\caption{Relationship between event-based F1-score and total context window duration across test sets. Each plot shows results for a model using equal context durations before and after a 16-second target segment.}
\label{fig:contextwindowlength}
\end{figure*}

Fig.~\ref{fig:contextwindowlength} shows how model performance varies with the total duration of the context window. Extending the context window generally leads to higher F1-scores, suggesting that incorporating segments surrounding the target segment helps the model better identify seizure-related patterns. The Siena dataset represents an exception, where increasing the window length shows no additional benefit, and the results exhibit considerable variance due to the limited number of seizures. For the other test sets, the performance gains seem to saturate as the context window lengthens, suggesting that extending the temporal context beyond 128 seconds offers little added value.

\subsection{Model Ensembling}
\label{sec:Ensembling}

Model ensembling can often reduce variance and improve overall performance by combining the predictions of multiple independently trained models. In our experiment, we constructed an ensemble of three LookAroundNet variants, each trained with a different temporal context alignment: 64 seconds before, 64 seconds after, or centered with 32 seconds on either side of the target segment. Predicted probabilities from the models were averaged to obtain the final output. Performance metrics for the ensemble model are provided in Table~\ref{tab:performance}, and F1-score comparisons to the individual configurations are shown in Fig.~\ref{fig:windowvariation}.

Ensembling the configurations resulted in state-of-the-art performance across all datasets, outperforming both the individual models and previously published methods. The ensemble achieved the highest event-based F1-scores and the lowest number of false positives for all test sets. However, the seizure detection rate of the ensemble model was generally lower than that of the individual models. This trade-off, however, can be adjusted by tuning the prediction threshold to favor sensitivity or specificity, depending on the application.

Interestingly, we observed that ensembling different context window configurations yielded larger improvements than simply broadening the temporal context in a single model, as in Fig.~\ref{fig:contextwindowlength}. This indicates an advantage of using multiple context window configurations over a single, extended window.

\subsection{Cost of Inference}
\label{sec:inference}

Fast inference is a critical aspect of seizure detection, as it can improve workflow efficiency, enable rapid clinical response, and support real-time monitoring. To evaluate the practicality of LookAroundNet for clinical deployment, we measured its inference performance on one hour of EEG data across several computing platforms. The results of these experiments are summarized in Table~\ref{tab:inference}.

\begin{table}[t]

\caption{Inference time breakdown across pipeline steps and hardware configurations. }
\label{tab:inference}
\centering
    \begin{tabularx}{\columnwidth}{Xrrrr}
        \toprule
        \multirow{2}{*}{Hardware} & \multicolumn{4}{c}{Processing Steps (secs)} \\
        \cmidrule(l){2-5}
        & Load Model & Load EEG & Pre-process & Inference \\
        \midrule
         M3 Pro & 0.03 & 0.51 & 1.09 & 11.10 \\
         \midrule
          DGX Spark & 0.05 & 0.48 & 1.87 & 3.09\\
         \midrule
         QRTX 5000 & 0.12 & 0.85 & 1.80 & 2.64\\
        \bottomrule
    \end{tabularx}

    \vspace{2pt}
\parbox{\columnwidth}{%
Results are reported in seconds and correspond to the processing of a 1-hour EDF file using a single model with a total context window size of 64 seconds. Model inference costs about 2 GFLOPs for every second of EEG processed, assuming the inference window advances in 2-second steps.
}

\end{table}

Both an earlier-generation workstation equipped with an Intel Xeon Gold 5218 processor and an NVIDIA Quadro RTX 5000 GPU, and the NVIDIA DGX Spark, a newer desktop-sized AI computer, completed the task in under 6 seconds. In comparison, a MacBook Pro with an M3 chip took less than 13 seconds. All systems were able to perform inference many times faster than real-time, showing that the model runs efficiently even on standard, general-purpose hardware. Several steps in the pipeline, such as data loading, pre-processing, and inference, can run in parallel to increase throughput. With multiple GPUs, inference can be distributed across devices or used to run multiple models at once, for example, in an ensemble setup.

\section{Conclusions}

Our evaluation shows that LookAroundNet delivers competitive, state-of-the-art performance on a range of out-of-sample EEG datasets. The model shows strong generalization across diverse clinical settings, demographic groups, and both routine and ambulatory recordings. Furthermore, the results indicate that LookAroundNet has potential for clinical use, offering strong seizure detection performance with relatively low computational requirements.

LookAroundNet demonstrates strong performance compared to prior work when trained solely on the same publicly available datasets used in previous studies. Incorporating data from the proprietary Kvikna dataset of ambulatory recordings further enhances performance, emphasizing the importance of ample and diverse training data in achieving generalization. In addition, extending the model’s temporal context provided an additional boost in detection accuracy, and ensembling models with different context window configurations achieved the best performance.

Although the proposed model demonstrates encouraging performance, several limitations remain. The detection rate on ambulatory data is still relatively low, and certain signal artifacts continue to generate an excessive number of false positives. Moreover, the use of a fixed-length target segment can limit the model’s capacity to capture very short seizure events. These challenges may be alleviated by refining post-processing methods, such as adjusting the prediction threshold, and by enhancing artifact management through targeted training or artifact removal techniques.

Performance varied considerably across datasets, highlighting the impact of differences in recording quality, labeling practices, and clinical context. This reinforces the ongoing challenge of developing seizure datasets with consistent annotations that also capture the diversity of recording environments and patient populations. Consequently, the proposed model may underperform on data types that are underrepresented in this study, such as neonatal recordings.

This study opens up multiple avenues for further research. A worthwhile next step is to better understand the model’s failure modes, including which artifacts cause false predictions and which seizures are often missed. In that context, exploring the model's internal behavior, such as feature extraction or attention mechanisms, may offer insights into which temporal segments or EEG channels the model relies on for its decisions~\cite{isaev2020attention}. This could reveal useful patterns in attention distribution and support improvements in both interpretability and model performance.

For clinical deployment, a lightweight and energy-efficient version of the model is essential to enable real-time seizure detection on edge devices, such as those used in wearable or home-based monitoring systems~\cite{ingolfsson2025robust}. Seamless integration with existing EEG platforms and hospital infrastructure is also crucial to support practical workflows, and facilitate clinical analysis. These steps are critical to closing the gap between research and clinical application. As this work suggests, the path to better seizure detection may begin by learning to look around, not only within the data but also toward the clinical environments and technologies that will bring such models to life.

\section*{Acknowledgment}
We thank the team at Kvikna for their expert assistance in EEG data management and their valuable feedback throughout this project.

\ifdefined\SIMPLE
  \section*{Funding}
  This project received funding from the Icelandic Centre for Research through the Technology Development Fund (grant 2321720-601).
\fi

\section*{Declaration of Competing Interests}
Þór Sverrisson was an employee of Kvikna Consulting ehf. during part of the study period, when Kvikna Consulting ehf. and Kvikna Medical ehf. shared common ownership. Kvikna Medical specializes in the design and development of neurodiagnostic products and solutions under the StratusEEG brand. Steinn Guðmundsson declares no competing interests.

\section*{Ethics Statement}
The EEG data used in this study had been collected previously with informed consent, anonymised prior to analysis, and processed in accordance with local data-protection regulations. The University of Iceland Science Ethics Committee reviewed the study and found it to be compliant with the scientific ethics rules of the University of Iceland (Ref. SHV2025-113).

\section*{Data Availability}
The EEG datasets used in this study include both public and proprietary data. Public datasets were obtained from research databases, either freely accessible or subject to data use agreements. Proprietary EEG data were provided by Kvikna Medical ehf. and cannot be made publicly available due to confidentiality agreements.

\section*{Code Availability}
Code supporting the findings of this study, including scripts for data pre-processing, model training, and evaluation is publicly available at \url{https://github.com/ths220/lookaroundnet}.

\printbibliography[heading=bibintoc]

\newpage
\appendix
\section{Dataset Details}
\label{app:dataset}

\begin{figure*}[t]
\centerline{\includegraphics[width=\textwidth]{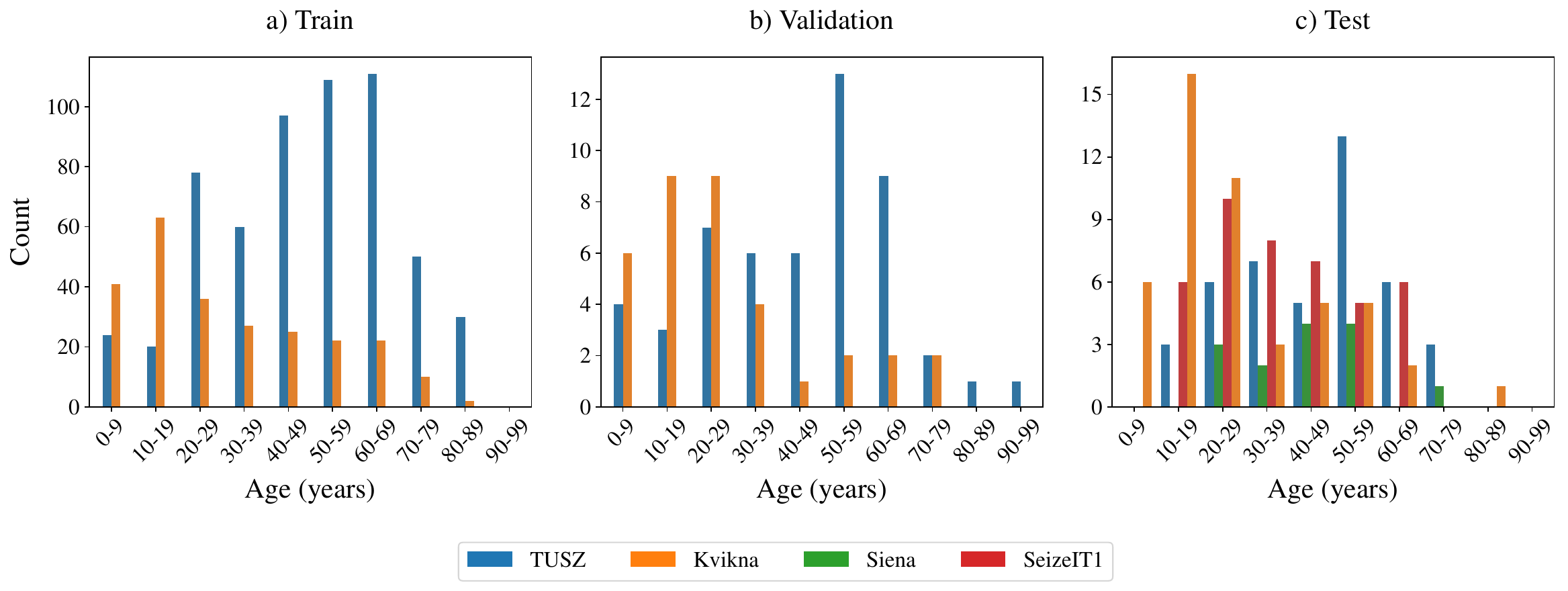}}
\caption{Age distribution of patients in the training, validation, and test sets.}
\label{fig:datasetage}
\end{figure*}

\begin{figure*}[t]
\centerline{\includegraphics[width=\textwidth]{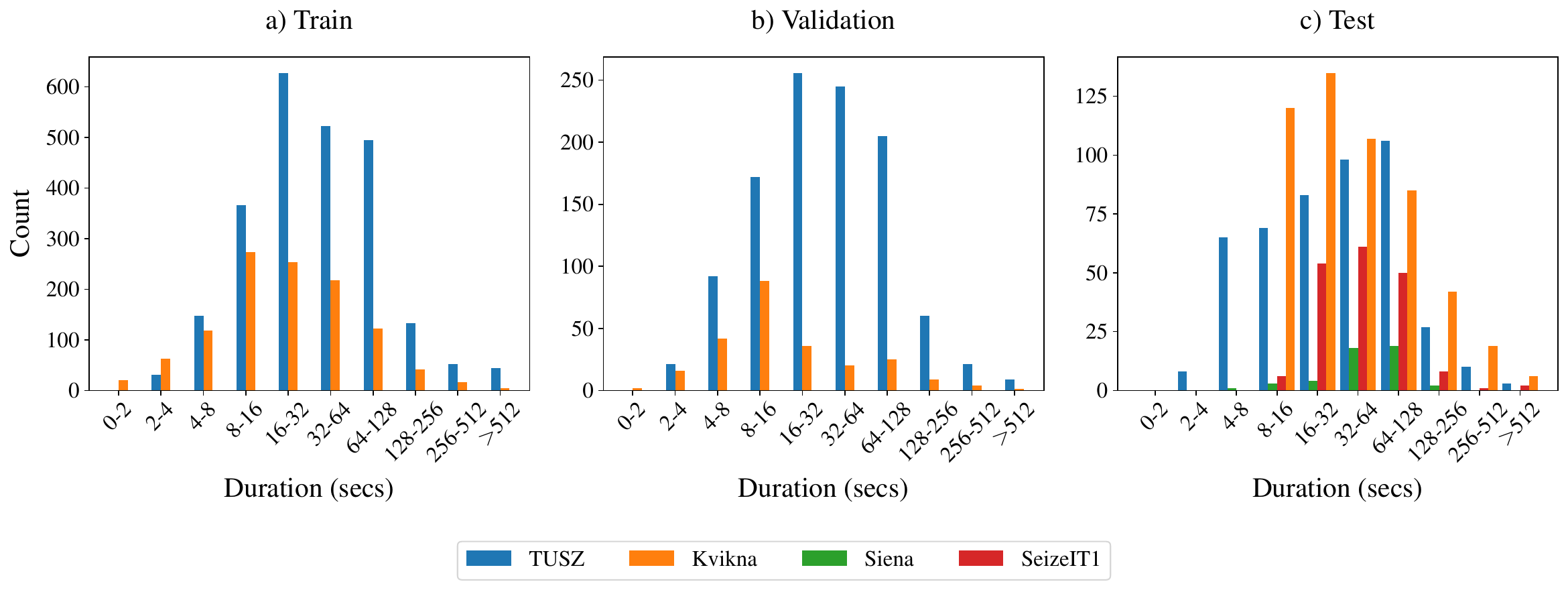}}
\caption{Distribution of seizures by duration across the training, validation, and test sets.}
\label{fig:datasetszdur}
\end{figure*}

Fig.~\ref{fig:datasetage} reports the age distribution of patients in the training, validation, and test sets. Fig.~\ref{fig:datasetszdur} shows the distribution of seizure durations across the same splits.

\section{Design Parameters}
\label{app:modelarchitecture}

\begin{table}[t]
\caption{Overview of the model parameters of the proposed LookAroundNet.}
\label{tab:lookaroundparameters}
\centering
\begin{tabular}{lr}
\toprule
Parameter & Value \\
\midrule
  Input Patch Size & 48\\
  Embedding Dim & 96\\
  Convolutional Layers & 2 \\
  Convolutional Filter (Kernel, Stride) & 3, 1\\
  Transformer Encoder Layers & 3\\ 
  Transformer Encoder Feedforward & 384\\
  Transformer Encoder Dropout & 0.1\\
  Attention Heads & 3\\ 
  Classifier Dropout & 0.5\\
\bottomrule
\end{tabular}

\end{table}

\begin{table}[t]
\caption{Overview of the training parameters.}
\label{tab:trainingparameters}
\centering
\begin{tabular}{lr}
\toprule
Parameter & Value \\
\midrule
  Optimizer & AdamW Schedule-Free \\
  Learning Rate & 5e-4 \\
  Batch Size & 512 \\
  Label Smoothing Factor & 0.1 \\
\bottomrule
\end{tabular}
\end{table}

Table~\ref{tab:lookaroundparameters} lists the hyperparameters of LookAroundNet. Table~\ref{tab:trainingparameters} summarizes the parameters used during training

\section{Summary of Results}
\label{app:completeresults}
\begin{table*}[t]
    \caption{Performance of LookAroundNet under SzCORE evaluation across context window configurations.}
    \label{tab:contextperformance}
    \centering
    \begin{tabularx}{\textwidth}{
    X
    l 
    *{4}{S[table-format=2.1(1)]}  % Event based numbers with uncertainty
    *{4}{S[table-format=2.1(1)]}  % Sample based numbers with uncertainty
    }

    \toprule
    \multirow{2}{*}{Context Window Range} &
    \multirow{2}{*}{Metrics} & 
    \multicolumn{4}{c}{Event based} & \multicolumn{4}{c}{Sample based} \\
    \cmidrule(lr){3-6} \cmidrule(lr){7-10}
     & & {TUSZ} & {Siena} & {SeizeIT} & {Kvikna} & {TUSZ} & {Siena} & {SeizeIT} & {Kvikna} \\
    \midrule
        \multirow{4}{*}{$-0, 0$}
             & F1-score       & 64.4 \pm 2.9 & 54.5 \pm 6.3  & 16.6 \pm 2.5 & 16.1 \pm 0.8 & 45.9 \pm 3.9 & 42.6 \pm 3.3 & 22.3 \pm 2.1 & 12.2 \pm 0.6 \\
              & Sensitivity & 58.2 \pm 2.9 & 55.9 \pm 5.7  & 37.1 \pm 3.3 & 35.2 \pm 2.4 & 33.0 \pm 4.1 & 32.0 \pm 3.3 & 20.3 \pm 1.0 & 12.4 \pm 0.7 \\
              & Precision   & 72.3 \pm 3.9 & 54.8 \pm 11.4 & 10.8 \pm 2.1 & 10.4 \pm 0.6 & 76.5 \pm 3.1 & 64.4 \pm 6.7 & 25.1 \pm 4.9 & 12.1 \pm 0.8 \\
              & FP/day      & 9.0 \pm 1.5  & 1.9 \pm 1.1   & 1.8 \pm 0.5  & 6.8 \pm 0.5  &              &              &              &              \\
            \midrule
            \multirow{4}{*}{$-8, 8$}
             & F1-score       & 69.3 \pm 1.2 & 58.4 \pm 4.8 & 20.0 \pm 1.5 & 18.3 \pm 2.4 & 53.4 \pm 1.1 & 48.6 \pm 2.4 & 23.0 \pm 1.8 & 13.3 \pm 1.9 \\
            & Sensitivity & 67.9 \pm 1.6 & 59.5 \pm 4.1 & 47.1 \pm 2.1 & 42.2 \pm 1.5 & 41.4 \pm 1.8 & 37.5 \pm 2.5 & 23.6 \pm 0.3 & 13.4 \pm 0.9 \\
              & Precision   & 70.8 \pm 2.2 & 58.0 \pm 8.9 & 12.7 \pm 1.2 & 11.7 \pm 2.0 & 75.8 \pm 3.9 & 69.2 \pm 4.5 & 22.7 \pm 3.9 & 13.7 \pm 3.8 \\
              & FP/day      & 11.2 \pm 1.3 & 1.7 \pm 0.7  & 1.9 \pm 0.2  & 7.4 \pm 1.6  &              &              &              &              \\
            \midrule
            \multirow{4}{*}{$-16, 16$}
             & F1-score       & 72.1 \pm 2.5 & 50.4 \pm 6.3 & 22.0 \pm 2.3 & 18.9 \pm 0.9 & 54.0 \pm 5.8 & 48.3 \pm 2.3 & 26.8 \pm 1.6 & 14.9 \pm 0.8 \\
              & Sensitivity & 71.0 \pm 5.5 & 65.0 \pm 3.0 & 52.5 \pm 2.8 & 44.2 \pm 2.7 & 41.4 \pm 6.4 & 40.6 \pm 2.3 & 25.5 \pm 0.7 & 14.9 \pm 1.3 \\
              & Precision   & 73.5 \pm 1.8 & 41.3 \pm 7.3 & 13.9 \pm 1.8 & 12.0 \pm 0.8 & 78.8 \pm 4.6 & 59.6 \pm 4.3 & 28.5 \pm 3.6 & 14.9 \pm 1.4 \\
              & FP/day      & 10.3 \pm 1.6 & 3.6 \pm 0.8  & 1.9 \pm 0.3  & 7.4 \pm 0.9  &              &              &              &              \\
            \midrule
            \multirow{4}{*}{$-32, 32^{(1)}$}
             & F1-score       & 72.1 \pm 1.0 & 54.9 \pm 5.1 & 26.4 \pm 2.4 & 21.8 \pm 0.6 & 62.1 \pm 2.8 & 49.0 \pm 1.6 & 31.0 \pm 1.8 & 17.6 \pm 0.6 \\
              & Sensitivity & 74.2 \pm 1.9 & 69.1 \pm 3.4 & 58.5 \pm 2.2 & 47.4 \pm 3.0 & 51.9 \pm 2.6 & 42.0 \pm 1.4 & 29.3 \pm 1.2 & 16.4 \pm 0.9 \\
              & Precision   & 70.3 \pm 3.0 & 46.0 \pm 7.2 & 17.1 \pm 2.1 & 14.2 \pm 0.6 & 77.4 \pm 3.1 & 59.4 \pm 6.5 & 33.0 \pm 3.5 & 18.9 \pm 1.0 \\
              & FP/day      & 12.7 \pm 2.1 & 3.2 \pm 1.0  & 1.7 \pm 0.3  & 6.5 \pm 0.7  &              &              &              &              \\
             \midrule
             \multirow{4}{*}{$-64, 64$}
             & F1-score       & 73.8 \pm 2.7 & 52.1 \pm 4.9 & 27.1 \pm 3.7 & 21.6 \pm 1.3 & 66.2 \pm 3.6 & 48.0 \pm 1.0 & 32.2 \pm 3.3 & 18.6 \pm 1.9 \\
              & Sensitivity & 73.2 \pm 1.5 & 75.9 \pm 1.2 & 58.5 \pm 4.8 & 50.9 \pm 2.7 & 54.9 \pm 3.4 & 42.7 \pm 2.0 & 29.5 \pm 2.4 & 19.1 \pm 1.6 \\
              & Precision   & 74.4 \pm 4.4 & 39.8 \pm 5.5 & 17.7 \pm 3.0 & 13.7 \pm 1.0 & 83.4 \pm 4.9 & 55.2 \pm 3.7 & 36.4 \pm 7.2 & 18.2 \pm 2.3 \\
              & FP/day      & 10.2 \pm 2.4 & 4.4 \pm 1.0  & 1.6 \pm 0.4  & 7.3 \pm 0.6  &              &              &              &              \\
            \midrule
            \multirow{4}{*}{$-64, 0^{(2)}$}
             & F1-score       & 69.5 \pm 1.0 & 59.4 \pm 1.7 & 24.5 \pm 2.6 & 23.1 \pm 1.4 & 59.6 \pm 4.6 & 49.0 \pm 2.3 & 29.0 \pm 1.8 & 18.0 \pm 2.0 \\
              & Sensitivity & 70.1 \pm 2.0 & 71.8 \pm 2.6 & 55.0 \pm 1.4 & 48.4 \pm 1.0 & 48.5 \pm 4.9 & 39.8 \pm 2.5 & 26.4 \pm 0.5 & 16.9 \pm 2.0 \\
              & Precision   & 68.8 \pm 1.4 & 50.8 \pm 2.5 & 15.8 \pm 2.2 & 15.2 \pm 1.3 & 77.6 \pm 4.4 & 63.8 \pm 3.2 & 32.5 \pm 5.3 & 19.8 \pm 3.8 \\
              & FP/day      & 12.8 \pm 1.0 & 2.6 \pm 0.3  & 1.7 \pm 0.3  & 6.2 \pm 0.7  &              &              &              &              \\
            \midrule
            \multirow{4}{*}{$-0, 64^{(3)}$}
             & F1-score       & 76.0 \pm 1.6 & 54.4 \pm 5.0 & 27.1 \pm 3.7 & 22.0 \pm 1.1 & 64.9 \pm 1.6 & 49.8 \pm 1.5 & 31.1 \pm 1.2 & 18.2 \pm 0.5 \\
              & Sensitivity & 72.8 \pm 1.3 & 65.5 \pm 3.7 & 51.9 \pm 0.8 & 44.5 \pm 1.8 & 52.4 \pm 2.0 & 41.4 \pm 1.5 & 27.3 \pm 1.4 & 16.9 \pm 1.4 \\
              & Precision   & 79.6 \pm 2.0 & 46.7 \pm 6.4 & 18.5 \pm 3.4 & 14.7 \pm 1.1 & 85.4 \pm 3.9 & 62.5 \pm 3.4 & 36.3 \pm 3.6 & 19.9 \pm 1.8 \\
              & FP/day      & 7.5 \pm 0.8  & 2.9 \pm 0.7  & 1.4 \pm 0.3  & 5.9 \pm 0.7  &              &              &              &              \\
             \midrule
             \multirow{4}{*}{Ensemble (1), (2), (3)}
              & F1-score       & 77.8 \pm 1.0 & 68.5 \pm 1.3 & 47.0 \pm 5.2  & 31.2 \pm 2.4 & 61.8 \pm 3.4 & 49.9 \pm 2.0 & 37.3 \pm 1.1 & 19.9 \pm 0.6 \\
              & Sensitivity  & 69.2 \pm 1.3 & 62.7 \pm 3.4 & 49.0 \pm 1.3  & 38.4 \pm 1.1 & 46.7 \pm 3.6 & 37.5 \pm 2.0 & 25.4 \pm 0.9 & 13.3 \pm 0.8 \\
              & Precision    & 89.0 \pm 1.5 & 75.6 \pm 2.3 & 46.2 \pm 10.4 & 26.6 \pm 3.8 & 91.6 \pm 3.5 & 74.5 \pm 0.8 & 71.0 \pm 8.2 & 40.2 \pm 6.  \\
              & FP/day       & 3.4 \pm 0.5  & 0.8 \pm 0.1  & 0.4 \pm 0.2   & 2.5 \pm 0.5  &              &              &              &              \\
              \bottomrule
    \end{tabularx}
    
    \vspace{2pt}
    \parbox{\textwidth}{%
    TUSZ and Kvikna datasets are utilized for training.}

\end{table*}

\begin{table*}[t]
    \caption{Performance of LookAroundNet under SzCORE evaluation for various training sets.}
    \label{tab:trainingsetperformance}
    \centering
    \begin{tabularx}{\textwidth}{
    X
    l 
    *{4}{S[table-format=2.1(1)]}  % Event based numbers with uncertainty
    *{4}{S[table-format=2.1(1)]}  % Sample based numbers with uncertainty
    }

    \toprule
    \multirow{2}{*}{Training set} &
    \multirow{2}{*}{Metrics} & 
    \multicolumn{4}{c}{Event based} & \multicolumn{4}{c}{Sample based} \\
    \cmidrule(lr){3-6} \cmidrule(lr){7-10}
     & & {TUSZ} & {Siena} & {SeizeIT} & {Kvikna} & {TUSZ} & {Siena} & {SeizeIT} & {Kvikna} \\
    \midrule
    \multirow{4}{*}{TUSZ}
     & F1-score       & 68.7 \pm 1.1 & 46.1 \pm 4.1 & 12.6 \pm 1.0 & 14.2 \pm 1.2 & 60.2 \pm 3.5 & 45.2 \pm 3.6 & 18.3 \pm 1.8 & 11.5 \pm 1.2 \\
      & Sensitivity & 73.3 \pm 2.1 & 63.2 \pm 6.3 & 48.6 \pm 3.1 & 36.5 \pm 1.4 & 49.6 \pm 3.7 & 37.9 \pm 3.7 & 23.8 \pm 1.8 & 12.2 \pm 0.6 \\
      & Precision   & 64.6 \pm 1.2 & 36.6 \pm 4.7 & 7.2 \pm 0.6  & 8.8 \pm 0.9  & 76.8 \pm 4.3 & 56.3 \pm 4.7 & 15.2 \pm 2.9 & 11.0 \pm 1.6 \\
      & FP/day      & 16.1 \pm 1.0 & 4.2 \pm 1.0  & 3.6 \pm 0.1  & 8.6 \pm 0.8  &              &              &              &              \\
     \midrule
    \multirow{4}{*}{TUSZ, Siena}
     & F1-score       & 66.2 \pm 0.8 &  & 8.8 \pm 1.0  & 12.6 \pm 2.5 & 61.9 \pm 1.8 &  & 14.6 \pm 1.6 & 11.9 \pm 2.0 \\
      & Sensitivity & 75.4 \pm 1.1 &  & 51.2 \pm 2.0 & 40.7 \pm 2.9 & 52.7 \pm 4.4 &  & 27.2 \pm 2.8 & 14.5 \pm 1.3 \\
      & Precision   & 59.0 \pm 1.6 &  & 4.8 \pm 0.6  & 7.5 \pm 1.7  & 75.5 \pm 4.2 &  & 10.1 \pm 1.7 & 10.1 \pm 2.2 \\
      & FP/day      & 21.0 \pm 1.6 &  & 5.9 \pm 0.5  & 11.6 \pm 1.7 &              &  &              &              \\
     \midrule
    \multirow{4}{*}{TUSZ, SeizeIT1}
     & F1-score       & 70.7 \pm 3.0 & 33.3 \pm 7.2 &  & 16.1 \pm 2.6 & 60.4 \pm 4.0 & 35.9 \pm 6.4  &  & 12.7 \pm 1.5 \\
      & Sensitivity & 73.0 \pm 1.3 & 73.6 \pm 5.5 &  & 39.9 \pm 2.6 & 49.5 \pm 4.1 & 44.3 \pm 3.8  &  & 13.5 \pm 1.2 \\
      & Precision   & 68.7 \pm 4.7 & 22.0 \pm 6.5 &  & 10.2 \pm 2.2 & 77.8 \pm 4.9 & 32.2 \pm 12.1 &  & 12.5 \pm 3.4 \\
      & FP/day      & 13.4 \pm 2.6 & 10.8 \pm 4.5 &  & 8.4 \pm 2.7  &              &               &  &              \\
    \bottomrule
    \end{tabularx}
    
    \vspace{2pt}
    \parbox{\textwidth}{%
    LookAroundNet uses a context window configuration of $(-32,32)$.}

\end{table*}

Table~\ref{tab:contextperformance} reports SZCORE results for LookAroundNet across different context window configurations, while Table~\ref{tab:trainingsetperformance} reports results for different training sets.

\section{Visualization of Model Outputs}
\label{app:visualoutput}

\begin{figure*}[!t]
\centerline{\includegraphics[width=\textwidth]{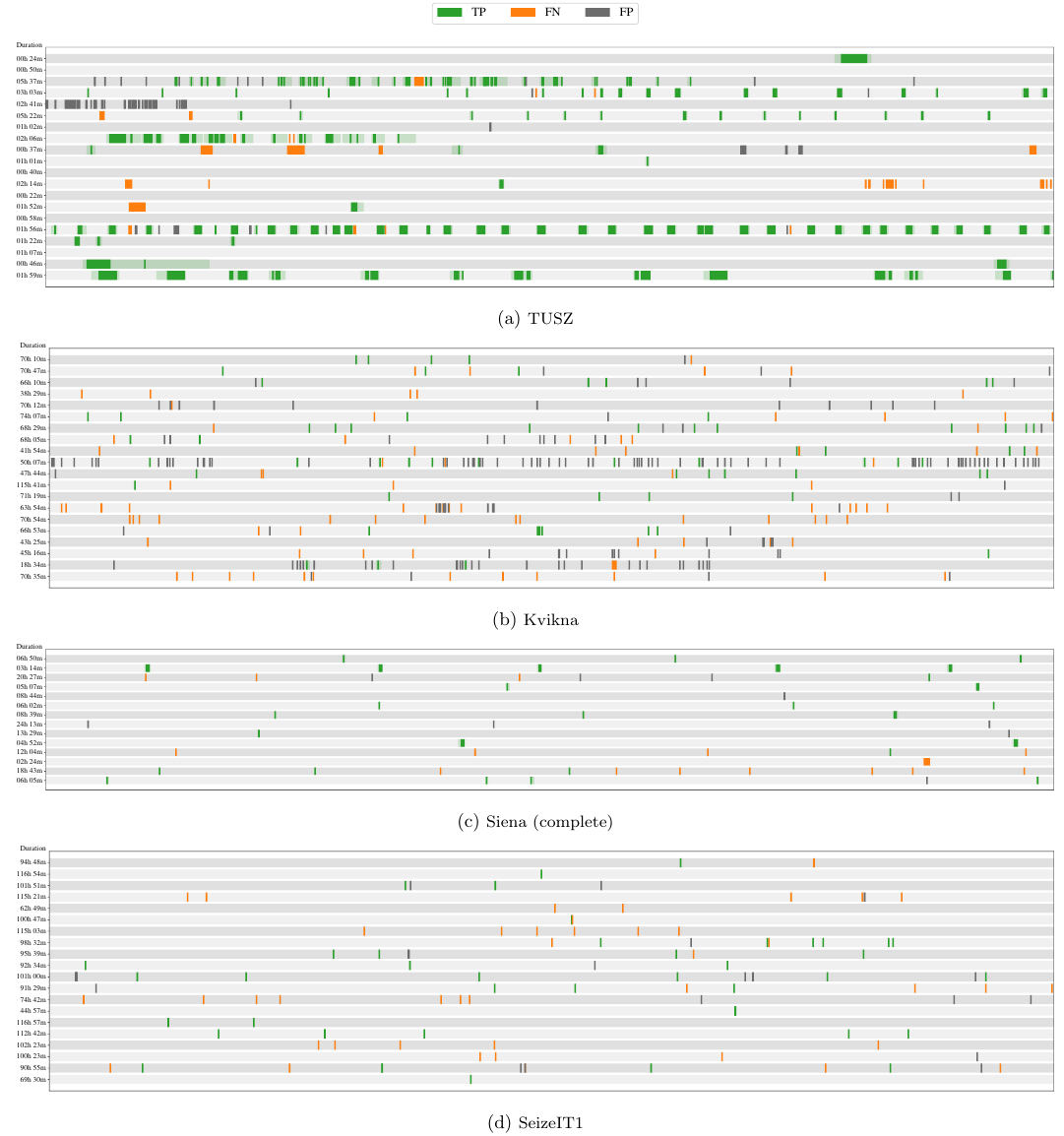}}

\caption{Examples of model predictions versus actual seizures for test set recordings using the ensemble model. Correctly predicted seizures are shown in green, with the predicted segment highlighted. Missed seizures are shown in orange, and false predictions in black. Each line represents one or more recordings from a single patient. The recordings may be discontinuous and are stretched to fit the plot width. As a result, equal-length segments on the plot may correspond to different actual durations across patients.}
\label{fig:model_pred}
\end{figure*}

Fig.~\ref{fig:model_pred} illustrates examples of ensemble model predictions compared to ground truth seizure annotations for test set recordings.

\end{document}